\documentclass[conference]{IEEEtran}
\IEEEoverridecommandlockouts
\usepackage{cite}
\usepackage{amsmath,amssymb,amsfonts}
\usepackage{algorithmic}
\usepackage{graphicx}
\usepackage{textcomp}
\usepackage{xcolor}
\def\BibTeX{{\rm B\kern-.05em{\sc i\kern-.025em b}\kern-.08em
    T\kern-.1667em\lower.7ex\hbox{E}\kern-.125emX}}
    
\usepackage{color}
\usepackage{tabularx}
\usepackage{multirow}
\usepackage{amsmath}
\usepackage{amssymb}
\usepackage{pifont}
\usepackage{hyperref}
\usepackage{colortbl}
\hypersetup{hypertex=true,
colorlinks=true,
linkcolor=blue,
anchorcolor=blue,
citecolor=blue}
\definecolor{mygray}{gray}{.9}

\begin{document}

\title{SCJD: Sparse Correlation and Joint Distillation for Efficient 3D Human Pose Estimation
}

\author{Weihong Chen$^1$, Xuemiao Xu$^{1\dagger}$, Haoxin Yang$^{1\dagger}$, Yi Xie$^1$, \\ Peng Xiao$^1$, Cheng Xu$^2$, Huaidong Zhang$^1$, Pheng-Ann Heng$^3$. \\
$^1$South China University of Technology.
$^2$The Hong Kong Polytechnic University. \\
$^3$ The Chinese University of Hong Kong.
\thanks{Accepted by ICME 2025 (Oral Paper).}
\thanks{The work is supported by China National Key R\&D Program (Grant No. 2023YFE0202700, 2024YFB4709200), Guangdong Provincial Natural Science Foundation for Outstanding Youth Team Project (No. 2024B1515040010), NSFC Key Project (No. U23A20391), Key-Area Research and Development Program of Guangzhou City (No. 2023B01J0022).}
\thanks{$^\dagger$Corresponding authors: Xuemiao Xu \& Haoxin Yang (email: xuemx@scut.edu.cn, harxis@outlook.com)}
}

\maketitle

\begin{abstract}
Existing 3D Human Pose Estimation (HPE) methods achieve high accuracy but suffer from computational overhead and slow inference, while knowledge distillation methods fail to address spatial relationships between joints and temporal correlations in multi-frame inputs. In this paper, we propose \textit{Sparse Correlation and Joint Distillation} (SCJD), a novel framework that balances efficiency and accuracy for 3D HPE. SCJD introduces \textit{Sparse Correlation Input Sequence Downsampling} to reduce redundancy in student network inputs while preserving inter-frame correlations. For effective knowledge transfer, we propose \textit{Dynamic Joint Spatial Attention Distillation}, which includes \textit{Dynamic Joint Embedding Distillation} to enhance the student's feature representation using the teacher's multi-frame context feature, and \textit{Adjacent Joint Attention Distillation} to improve the student network's focus on adjacent joint relationships for better spatial understanding. Additionally, \textit{Temporal Consistency Distillation} aligns the temporal correlations between teacher and student networks through upsampling and global supervision. Extensive experiments demonstrate that SCJD achieves state-of-the-art performance. Code is available at \url{https://github.com/wileychan/SCJD}.
\end{abstract}

\begin{IEEEkeywords}
3D Human Pose Estimation, distillation.
\end{IEEEkeywords}
\section{Introduction}
3D Human Pose Estimation (HPE) detects 3D joint coordinates from monocular video and is widely used in downstream tasks like action recognition~\cite{Rajasegaran2023cvpr, lin2024mutual} and motion analysis~\cite{yu2023Toward, xiaopeng_eccv}. To achieve high accuracy, prior methods typically use multi-frame input sequences to capture temporal dependencies~\cite{eccv_repose, Peng_2024_CVPR, Tang_2023_CVPR,ijcai23hdformer, zhang2022mixste, li2022mhformer, xue2022boosting} and large networks (e.g., Vision Transformers~\cite{dosovitskiy2021image}) for feature extraction. However, these approaches are computationally expensive and suffer from slow inference speeds due to dense inputs and significant network overhead.

To balance 3D HPE computational efficiency and accuracy, recent studies have adopted strided convolutions~\cite{li2022exploiting} or frequency-domain features~\cite{zhao2023poseformerv2,tang2023ftcm} to avoid processing long input sequences. However, these approaches require deeper and wider networks to sustain accuracy, still resulting in significant computational demands.
Knowledge distillation (KD)~\cite{hinton2015distilling} has also been explored to address this challenge~\cite{hwang2020lightweight,kim2024pruning,bian2024learning}. These methods train a light student network to replicate the output logits~\cite{hwang2020lightweight} or the final-layer features~\cite{kim2024pruning,bian2024learning} of heavy teacher network. Despite their potential, these methods are limited by their reliance on single-image inputs, neglecting the temporal correlations between actions in consecutive frames—an essential factor in resolving the 2D-to-3D ambiguity in monocular HPE.
Furthermore, these approaches fail to emphasize that the student network should prioritize learning the relationships between human joints from the teacher network rather than merely replicating final outputs or high-level semantic features. These limitations ultimately restrict their accuracy and performance.

\begin{figure}
    \centerline{\includegraphics[width=\columnwidth]{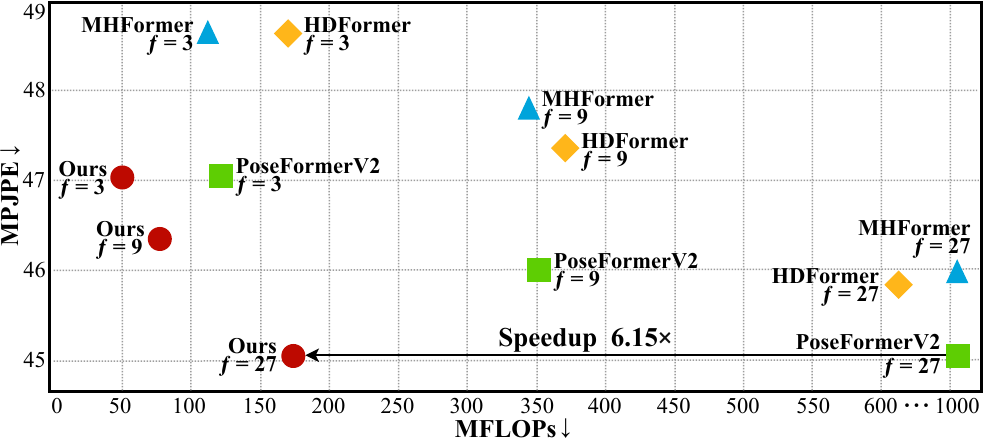}} 
    \setlength{\abovecaptionskip}{0.05cm}
    \caption{Comparison between the previous methods PoseFormerV2~\cite{zhao2023poseformerv2}, HDFormer~\cite{ijcai23hdformer}, MHFormer~\cite{li2022mhformer} and our SCJD on Human3.6M. $f$ represents the length of the input sequence. }
    \label{intro}
    \vspace{-20pt}
\end{figure}


To address these challenges, we propose a novel \textit{\textbf{S}}parse \textit{\textbf{C}}orrelation and \textit{\textbf{J}}oint \textit{\textbf{D}}istillation (SCJD) for 3D HPE. 
Our key insight is to leverage sparse input sequences and KD to reduce computational overhead while maintaining accuracy. Specifically, Our approach begins with \textit{Sparse Correlation Input Sequence Downsampling}, which sparsifies the student network's inputs to reduce redundancy while preserving inter-frame correlations more effectively than single-frame methods.
Next, to transfer knowledge effectively from a dense, heavy teacher network to a sparse, lightweight student network, we introduce \textit{Dynamic Joint Spatial Attention Distillation}, which comprises two components: \textit{a) Dynamic Joint Embedding Distillation}, which aggregates the local context information of teacher's spatial encoder to enhance the student's spatial feature representation. \textit{b) Adjacent Joint Attention Distillation}, which employs an adjacent joint attention matrix to guide the student network in focusing on connections between adjacent joints, improving its understanding of human posture.
Finally, we propose \textit{Temporal Consistency Distillation}, which aligns the temporal correlation between input contexts by upsampling the student's features to match the teacher network's outputs, and complemented by global supervised learning with ground truth.
Our extensive experiments validate our method’s superiority over existing state-of-the-art techniques in terms of both efficiency and accuracy. Our contributions are threefold:
\begin{itemize}
    \item We propose a novel framework SCJD, which enhances the efficiency and accuracy of 3D HPE by utilizing sparse correlation inputs and dynamic joint spatial attention distillation.
    \item We introduce dynamic joint embedding distillation and adjacent joint attention distillation, enabling better alignment of joint spatial features between the teacher and student spatial encoder, significantly improving the student model's performance.
    \item Extensive experiments demonstrate that our method achieves superior speed-accuracy trade-offs, delivering up to 6.15$\times$ speedup with comparable accuracy, as illustrated in Fig.~\ref{intro}.
\end{itemize}

\begin{figure*}
    \centering
    \centerline{\includegraphics[width=\linewidth]{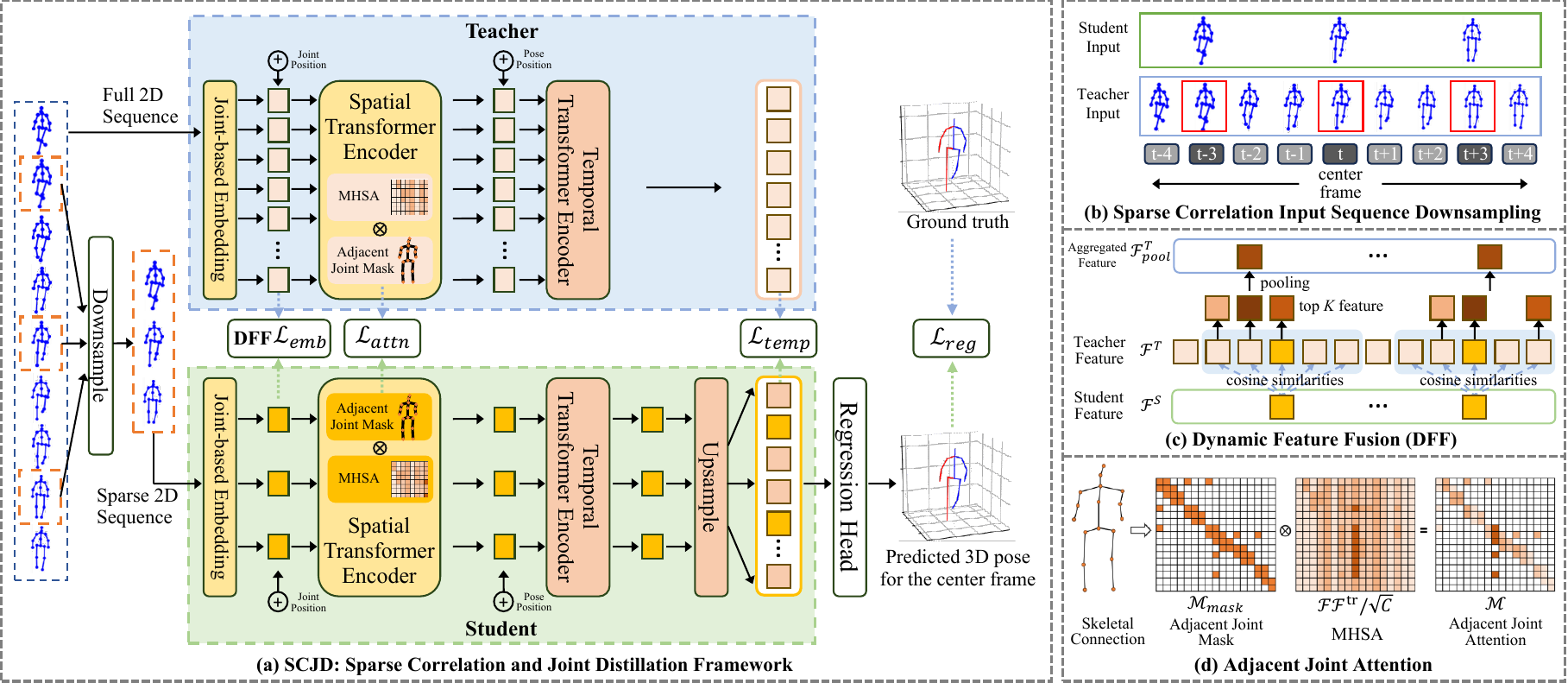}}
    
    \caption{Overview of the SCJD framework. SCJD comprises two transformer encoders that separately learn spatial and temporal information. The student model first employs \textit{Sparse Correlation Input Sequence Downsampling} to reduce input redundancy while preserving temporal correlations. Then, through \textit{Dynamic Joint Embedding Distillation} and \textit{Adjacent Joint Attention Distillation}, knowledge from the teacher network's spatial encoder is transferred to help the student model learn joint associations. Finally, \textit{Temporal Consistency Distillation} facilitates the transfer of temporal knowledge from the teacher's temporal encoder, enabling the student model to capture the temporal relationships between multi-frame postures.}
    \label{framework}
\vspace{-15pt}
\end{figure*}
\section{Related Work}

\subsection{Efficient 3D Human Pose Estimation}  
The goal of efficient 3D human pose estimation is to strike a balance between accuracy and computational efficiency. While recent studies~\cite{eccv_repose, Peng_2024_CVPR, Tang_2023_CVPR, ijcai23hdformer, zhang2022mixste, li2022mhformer, xue2022boosting} have prioritized improving accuracy, fewer have tackled the challenge of computational efficiency. Early efficient methods, such as Choi et al.~\cite{choi2021mobilehumanpose}, achieved faster pose estimation by utilizing compact networks and skip concatenation to reduce computational complexity. However, these approaches rely on single-frame inputs, which lack temporal context and limit accuracy compared to more advanced methods. Multi-frame inputs improve accuracy but increase computational demands. To address this, Li et al.~\cite{li2022exploiting} used strided convolutions to reduce the sequence length while capturing local context. Zhao et al.~\cite{zhao2023poseformerv2} encoded longer sequences into low-frequency coefficients combined with shorter pose sequences. While these methods reduce input length, they still require deeper, wider networks, leading to higher computational costs.

\subsection{Knowledge Distillation in 3D Human Pose Estimation}  
Knowledge distillation~\cite{hinton2015distilling} is an effective technique for compressing networks by transferring knowledge from a large teacher network to a lightweight student network, improving the latter's accuracy. Several studies~\cite{hwang2020lightweight, kim2024pruning, bian2024learning} have applied distillation to accelerate 3D human pose estimation. Hwang et al.~\cite{hwang2020lightweight} trained a student network to mimic the teacher's output logits, while Kim et al.~\cite{kim2024pruning} used pruning and feature imitation to boost performance. Bian et al.~\cite{bian2024learning} transferred multi-view stereo knowledge to the student model. However, these methods are limited to single-frame inputs and fail to utilize multi-frame video context, which is crucial for resolving 2D-to-3D ambiguities. Additionally, they overlook the importance of teaching the student network to learn joint spatial relationships. Our approach addresses these gaps by designing a distillation framework that incorporates multi-frame context while emphasizing joint relationships.

\section{Proposed Method}

\subsection{Overview} 
To maintain 3D HPE performance while maximizing computational efficiency, we propose SCJD, a framework that seamlessly integrates sparse correlation inputs and dynamic joint distillation for efficient and accurate 3D HPE. The detailed network architecture is illustrated in Fig.~\ref{framework}(a). 
Building on PoseFormer~\cite{zheng20213d}, we introduce \textit{Sparse Correlation Input Sequence Downsampling} to reduce input redundancy while preserving temporal correlations. Additionally, we employ \textit{Dynamic Joint Embedding Distillation}, \textit{Adjacent Joint Attention Distillation}, and \textit{Temporal Consistency Distillation} to transfer knowledge from the teacher network to a lightweight student model. This approach enables the network to efficiently process a 2D pose input sequence \(X \in \mathbb{R}^{f \times (J \cdot 2)}\) of \(f\) frames, where \(J\) represents the number of joints, and accurately predict the 3D pose \(y \in \mathbb{R}^{1 \times (J \cdot 3)}\) for the center frame of the sequence.


\subsection{Sparse Correlation Input Sequence Downsampling}
To minimize redundant information across video frames, we propose leveraging sparse inputs to expand the receptive field, enhancing both model accuracy and efficiency. Unlike the teacher network, which uses continuous 2D pose sequences, our lightweight student network employs \textit{sparse correlation input sequences} through strategic sampling. 
As illustrated in Fig.~\ref{framework}(b), for a 2D pose sequence of length \(f^T\) in the teacher network, the student network samples equidistantly both forward and backward from the center frame using a stride \(L\), ensuring alignment between the student and teacher networks at the center frame. This sampling reduces the student network's input sequence length to \(f^S = f^T / L\). For example, if the teacher network processes a sequence of \(f^T = 81\) frames, setting the stride \(L\) to \(\{3, 9, 27\}\) yields corresponding sparse sequence lengths of \(f^S = \{27, 9, 3\}\) for the student network. 
This approach drastically lowers the computational cost for the student network while maintaining the same large receptive field as the teacher network, ensuring an effective balance between efficiency and performance.

\subsection{Dynamic Joint Spatial Attention Distillation}
To capture the spatial relationships between human joints, we first utilize a spatial transformer to model the positional connections among joints in a frame of pose.  
For preprocessing the 2D pose sequence as network input, joint coordinates are linearly projected into a higher-dimensional space, and a learnable spatial positional embedding is added to encode the positions of the joints.  
Consequently, the joint embedding for each frame in the student and teacher networks are represented as \(\mathcal{F}^S \in \mathbb{R}^{1 \times J \times C^S}\) and \(\mathcal{F}^T \in \mathbb{R}^{1 \times J \times C^T}\), respectively, where \(C^S\) and \(C^T\) denote the feature dimensions of the joint-based embeddings. These features are then passed through a spatial transformer encoder. To minimize the model size, the student network utilizes a reduced feature dimension (\(C^S\)) compared to the teacher network (\(C^T\)).

\textbf{Dynamic Joint Embedding Distillation.}   
Our framework employs a sequence-to-frame approach, utilizing multi-frame 2D pose inputs to predict the 3D human pose of the central frame. To reduce computation, we apply sparse correlation input downsampling, which retains some degree of correlation while inevitably losing certain useful information due to the downsampling process. To address this limitation, we introduce a dynamic feature fusion (DFF) strategy, allowing the student network to learn multi-frame joint embedding features from the teacher network and improve its representation.

Specifically, in the early training stages (first \textit{N} epochs), when the student network has not yet learned meaningful knowledge, we simply use evenly spaced intervals to align the teacher sequence with the student's downsampling sequence. In later stages, as the student begins to acquire knowledge from the teacher, we leverage the multi-frame features of the teacher to help the student emulate the teacher's contextual feature.  
As shown in Fig.~\ref{framework}(c), for each pose \(p\) in the student network's sampled sequence, we locate the corresponding pose \(p\) in the teacher network's sequence. Centered at \(p\), we extract a feature sequence \( \left\{\mathcal{F}_k^T | k \in [p-(L-1), p+(L-1)]\right\}\) based on a predefined stride \textit{L}. We then compute cosine similarities between each teacher feature \(\mathcal{F}_k^T\) and the student feature \(\mathcal{F}_p^S\). The top-\(K\) teacher features with the highest similarity are pooled, producing \(\mathcal{F}_{pool}^T\). Finally, we minimize the distance between \(\mathcal{F}_{pool}^T\) and \(\mathcal{F}_p^S\), enabling the student to learn the teacher's contextual relationships effectively.
For the above two stages, the distillation loss of the joint-based embedding feature
can be formulated as:
\begin{equation}
    \mathcal{L}_{emb} = 
     \begin{cases}
    \frac{1}{J} \sum_{i=1}^{J} {\Vert \mathrm{FC}(\mathcal{F}_{p,i}^S) - \mathcal{F}_{p,i}^T \Vert }_2, &\text{{\scriptsize epoch}} \leq N, \\
      \frac{1}{J} \sum_{i=1}^{J} {\Vert \mathrm{FC}(\mathcal{F}_{p,i}^S) - \mathcal{F}_{pool,i}^T) \Vert }_2, &\text{{\scriptsize epoch}} > N,  
    \end{cases}
\end{equation}
where \(\mathrm{FC}(\cdot)\) is a linear layer to align the joint-based embedding dimensions of the student's \(C^S\) and the teacher's \(C^T\).
Through dynamic joint embedding distillation, the student network can learn more effective context representations.

\textbf{Adjacent Joint Attention Distillation.} 
After aligning the joint embedding features of the teacher and student networks, we initiate knowledge transfer. However, merely replicating the teacher network's feature extraction process is insufficient for the student network to grasp the underlying relationships between human joints.  
Human posture is naturally represented as a skeletal model, where interconnected joints form a structured topology, as illustrated in Fig.~\ref{framework}(d). In this structure, adjacent joints are directly connected keypoints, capturing local spatial relationships crucial to human pose representation. These interactions encode essential skeletal characteristics and provide valuable spatial-temporal constraints and biomechanical priors for the model.  
The connectivity of adjacent joints serves as critical prior knowledge, guiding the student network to focus on these relationships and enhancing its understanding of human posture. To incorporate this structural prior, we introduce an adjacent joint mask \(\mathcal{M}_{mask}\), which directs the student network to learn the teacher network’s patterns between adjacent joints. This \(\mathcal{M}_{mask}\) is defined as:

\begin{equation}
 \setlength{\abovedisplayskip}{-8pt}
    \mathcal{M}_{mask}(i, j) = \begin{cases} 
    1, &\text{if bone }i\text{ is connected to bone }j, \\
    0, &\text{otherwise}.
    \end{cases}
    \vspace{-8pt}
\end{equation}

To encourage the student network to focus on adjacent joint features derived from the teacher network, we multiply \(\mathcal{M}_{mask}\) with the attention matrix generated by both teacher and student network's spatial encoder. This results in the adjacent joint attention \(\mathcal{M}\), which filters out irrelevant elements in the attention matrix. The computation is expressed as:  
\begin{equation}
    \mathcal{M}(\mathcal{F}) = \frac{\mathcal{F} \mathcal{F}^{tr}}{\sqrt{C}} \otimes \mathcal{M}_{mask},
    \vspace{-5pt}
\end{equation}
where \( \mathcal{F} \in \mathbb{R}^{J \times C} \) represents the features from the spatial encoder of either the student or teacher network, \(J\) is the total number of joints, \(C\) is the embedding dimension, and \(\mathcal{F}^{tr}\) denotes the transpose of the feature matrix.  
To align skeletal relationships, we propose an adjacent joint attention distillation method that minimizes the difference between the skeleton correlation matrices of the teacher and student networks. This is formulated as:  
\begin{equation}  
\setlength{\abovedisplayskip}{0.2cm}
    \mathcal{L}_{attn} = \sum\nolimits_{i=1}^{J} \sum\nolimits_{j=1}^{J} \|\mathcal{M}_{i,j}(\mathcal{F}^S) - \mathcal{M}_{i,j}(\mathcal{F}^T)\|_2.  
\end{equation}  
This loss function encourages the student spatial encoder to better capture the relationships between adjacent joints, thereby improving its ability to understand the underlying skeletal structure for 3D human pose estimation.


\subsection{Temporal Consistency Distillation}
To enhance the smoothness of 3D HPE predictions across consecutive frames, we further process the temporal relationships between frames after capturing the spatial relationships between joints.  
For input sequences of lengths \(f^T\) (teacher) and \(f^S\) (student), the output features from their respective spatial transformer encoders are represented as \(Z^T \in \mathbb{R}^{f^T \times (J \cdot C^T)}\) and \(Z^S \in \mathbb{R}^{f^S \times (J \cdot C^S)}\). To retain frame temporal position information, a learnable temporal positional embedding is added to \(Z^T\) and \(Z^S\).  
These features, enriched with temporal positional embeddings, are then passed through their respective temporal transformer encoders to capture frame-to-frame dependencies across the sequence. The outputs are denoted as \(Y^T \in \mathbb{R}^{f^T \times (J \cdot C^T)}\) and \(Y^S \in \mathbb{R}^{f^S \times (J \cdot C^S)}\), corresponding to the teacher and student networks, respectively.

To generate dense 3D poses for all frames within the receptive field, we upsample the student network's final-layer features using a 1D deconvolution operation \(\textit{f}_{\rm deconv}\) to match the frame length of the teacher network. A fully connected layer is then applied to align the feature dimensions \(C^S\) and \(C^T\). The recovered sequence is expressed as:  
\begin{equation}
\setlength{\abovedisplayskip}{0.2cm}
    \hat{Y}^{S} = {\rm FC}(\textit{f}_{\rm deconv}(Y^S)),
\end{equation}
where \(\hat{Y}^S \in \mathbb{R}^{f^T \times (J \cdot C^T)}\). This step enables the transfer of comprehensive information from the teacher network's receptive field to the student network via token-level feature distillation. The temporal distillation process is formulated as:  
\begin{equation}
\setlength{\abovedisplayskip}{0.2cm}
    \mathcal{L}_{temp} = \frac{1}{f^T} \sum\nolimits_{i=1}^{f^T}  {\Vert \hat{Y}^{S}_i - Y^T_i  \Vert }_2 .
\end{equation}

To further enhance prediction accuracy, we employ the Mean Per Joint Position Error (MPJPE) loss~\cite{ionescu2013human3} to regress the error between the predicted results and the ground truth.  
Since the recovered sequence \(\hat{Y}^S \in \mathbb{R}^{f^T \times (J \cdot C^T)}\) is used to predict the 3D pose of the central frame, we apply a 1D convolution to aggregate temporal information, producing an output \(\hat{y} \in \mathbb{R}^{1 \times (J \cdot C^T)}\). This output is then passed through a linear projection to map \(\hat{y}\) into the pose representation \(\hat{y} \in \mathbb{R}^{1 \times (J \cdot 3)}\).  
The regression loss can be expressed as:  
\begin{equation}
\setlength{\abovedisplayskip}{0.2cm}
    \mathcal{L}_{reg} = \frac{1}{J} \sum\nolimits_{i=1}^{J}  {\Vert \hat{y}_i - y_i \Vert }_2 ,
\end{equation}
where \(\hat{y}_i\) and \(y_i\) represent the predicted 3D joint position and the ground truth respectively. 

Thus, the total loss $\mathcal{L}_{total}$ of our proposed SCJD is:
\begin{equation}
\setlength{\abovedisplayskip}{0.2cm}
    \mathcal{L}_{total} =  \mathcal{L}_{reg} + \alpha \mathcal{L}_{emb} + \beta \mathcal{L}_{attn} + \gamma \mathcal{L}_{temp},
    \label{total_loss}
\end{equation}
where $\alpha$, $\beta$, and $\gamma$ are the weights of each component of the loss, respectively.
\section{Experiments}

\subsection{Experimental Settings}

\textbf{Datasets and Metrics.} We evaluated our model on two datasets, namely Human3.6M~\cite{ionescu2013human3} and MPI-INF-3DHP~\cite{mehta2017monocular}. For Human3.6M~\cite{ionescu2013human3}, we compute Mean Per Joint Position Error (MPJPE), which is the average euclidean distance in millimeters between the predicted and the ground-truth 3D joint coordinates. To evaluate the efficiency, we use FLOating-Point operations (FLOPs) to measure the computational cost. For MPI-INF-3DHP~\cite{mehta2017monocular}, we additionally compute the Percentage of Correct Keypoints (PCK) within a range of 150 mm, and Area Under Curve (AUC) for a range of PCK thresholds. 

\textbf{Implementation Details.} The input sequence lengths of the student network are \(f^S= \left\{3,9,27\right\}\). For the teacher network, we chose \(f^T = 81\). The embedding dimensions of the student and teacher network are \(C^S=16\) and \(C^T=32\). The number of joints \(J=17\). In dynamic token matching, we set \(N=20\), \(K=3\). We empirically set the hyperparameters of Eq.~\ref{total_loss} to \(\alpha = \beta = 1\), \(\gamma = 0.01\). We trained our model for 50 epochs using the Adam optimizer~\cite{yang2024g,yang2022individual,kang2025sita
} with learning rate of 1e-3 and weight decay of 0.1. For the network structure and other more details, please refer to the supplementary material.

\begin{table} [t!]
    \setlength{\tabcolsep}{3.5pt}
    \caption{Comparisons with Previous Methods on Human3.6M. The 2D poses obtained by CPN~\cite{chen2018cascaded} are used as inputs. Best in red, second best in blue.}
    \vspace{-8pt}
    \label{table1}
    \centering
    \begin{tabularx}{\columnwidth}{lcrrrr}
        \hline
        Method & Publication & \(f\) & Params(M) & MFLOPs↓ & MPJPE↓ \\
        \hline
        MoVNect~\cite{hwang2020lightweight} &WACV'20& 1 & 1.03 & 1.35 & 97.3\\
        MobileHuman~\cite{choi2021mobilehumanpose} &CVPRW'21& 1 & 2.24 & 3920 & 56.9\\
        \hline
        PoseFormer~\cite{zheng20213d}  & ICCV'21 & 3 & 9.58 & \textcolor{blue}{60.4} & 51.8 \\ 
        UAU~\cite{einfalt2023uplift} & WACV'23 &3 & 10.36 & 535 & 49.9\\
        MixSTE~\cite{zhang2022mixste} & CVPR'22 & 3 & 33.70 & 3420 & 49.6\\
        MHFormer~\cite{li2022mhformer}  & CVPR'22  & 3 & 18.92 & 114.4 & 48.7 \\ 
        HDFormer~\cite{ijcai23hdformer} & IJCAI'23 & 3 & \textcolor{blue}{3.70} & 163 & 48.7\\
        STCFormer~\cite{Tang_2023_CVPR}& CVPR'23 & 3 & 4.75 & 484.1 & 48.5\\
        KTPFormer~\cite{Peng_2024_CVPR} & CVPR'24 & 3 & 34.73 & 3540 & 47.3\\
        RePOSE~\cite{eccv_repose}& ECCV'24 & 3 & 15.94 & 1618.4 & 47.3\\
        PoseFormerV2~\cite{zhao2023poseformerv2} & CVPR'23 & 3 & 14.36 & 117.3 & \textcolor{blue}{47.1} \\ 
        \rowcolor{mygray}
        \textbf{SCJD (Ours)}  & & 3 & \textcolor{red}{2.79} & \textcolor{red}{51.2} & \textcolor{red}{47.0} \\
        \hline
        PoseFormer~\cite{zheng20213d}  & ICCV'21 & 9 & 9.58 & \textcolor{blue}{150} & 49.9 \\ 
        PAA~\cite{xue2022boosting} & TIP'22 & 9 & 6.21 & 1139.1 & 48.9 \\
        UAU~\cite{einfalt2023uplift}  & WACV'23 & 9  & 10.36 & 543 & 47.9 \\ 
        MHFormer~\cite{li2022mhformer}  & CVPR'22  & 9 & 18.92 & 342.9 & 47.8 \\
        HDFormer~\cite{ijcai23hdformer} & IJCAI'23 & 9 & \textcolor{blue}{3.70} & 367.2 & 47.3\\
        KTPFormer~\cite{Peng_2024_CVPR} & CVPR'24 & 9 & 34.74 & 10610 & 46.6\\
        RePOSE~\cite{eccv_repose}& ECCV'24 & 9 & 15.94 & 4860 & 46.5\\
        PoseFormerV2~\cite{zhao2023poseformerv2}  & CVPR'23 & 9  & 14.36 & 351.7 & \textcolor{red}{46.0} \\ 
        \rowcolor{mygray}
        \textbf{SCJD (Ours)} &  & 9 & \textcolor{red}{2.79} & \textcolor{red}{81.3} & \textcolor{blue}{46.3} \\
        \hline
        UAU~\cite{einfalt2023uplift} & WACV'23 &27 & 10.36 & 564 & 47.4\\
        PoseFormer~\cite{zheng20213d} & ICCV'21  & 27 & 9.59 & \textcolor{blue}{452} & 47.0 \\
        PAA~\cite{xue2022boosting} & TIP'22 & 27 & 6.21 & 1229.5 & 46.8 \\
        MHFormer~\cite{li2022mhformer} & CVPR'22   & 27 & 18.92 & 1031.8 & 45.9 \\
        HDFormer~\cite{ijcai23hdformer} & IJCAI'23 & 27 & \textcolor{blue}{3.70} & 652.2 & 45.8\\
        KTPFormer~\cite{Peng_2024_CVPR} & CVPR'24 & 27 & 34.77 & 31840 & 45.4\\
        PoseFormerV2~\cite{zhao2023poseformerv2} & CVPR'23 & 27 & 14.36 & 1054.8 & \textcolor{blue}{45.2}\\
        RePOSE~\cite{eccv_repose}& ECCV'24 & 27 & 15.95 & 14560 & \textcolor{red}{44.1}\\
        \rowcolor{mygray}
        \textbf{SCJD (Ours)}& & 27 & \textcolor{red}{2.80} & \textcolor{red}{171.5} & \textcolor{blue}{45.2} \\
        \hline
    \end{tabularx}
    \label{tab1}
    \vspace{-10pt}
\end{table}

\begin{figure}[t!]
    \centerline{\includegraphics[width=\columnwidth]{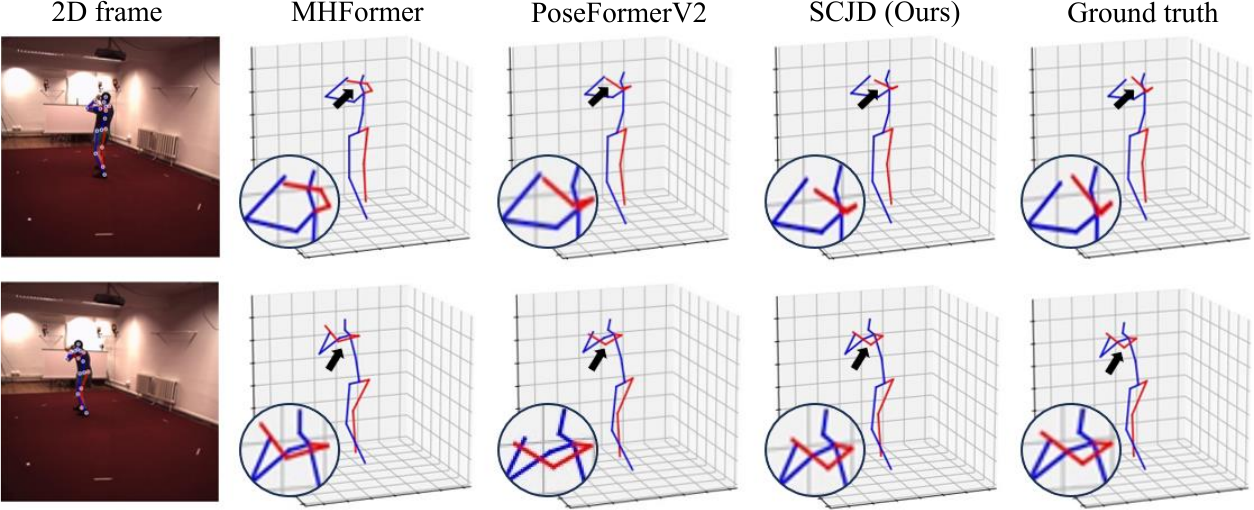}}
    \setlength{\abovecaptionskip}{0.05cm}
    \caption{Comparison of visualization results between ours and MHFormer~\cite{li2022mhformer} and PoseFormerV2~\cite{zhao2023poseformerv2} on Human3.6M test set S11 with the \textit{Photo} action. The black arrows highlight locations where our method has better results.}
    \label{exp-qualitative}
    \vspace{-15pt}
\end{figure}

\subsection{Comparisons with State-of-the-art Methods}
We quantitatively compare our student network's result with previous 3D HPE methods on Human3.6M~\cite{ionescu2013human3}, as shown in Table~\ref{table1}, where \(f\) denotes the input sequence's length. For each \(f\), we have achieved the minimum number of parameters, the minimum computational cost, and the optimal or suboptimal accuracy. Notably, when \(f\) is 27, with the same accuracy as PoseformerV2~\cite{zhao2023poseformerv2}, we have improved the speed by 6.15 times. When \(f\) is 3, with the similar computational cost to PoseFormer~\cite{zheng20213d}, we have improved the accuracy by 4.8mm MPJPE. 
Compared with the single-frame input methods~\cite{hwang2020lightweight,choi2021mobilehumanpose}, our SCJD significantly improves the accuracy, which proves that multi-frame input can effectively alleviate the ambiguity of 3D HPE.
Our SCJD is the only one that balances computational cost and accuracy. It proves that the student network benefits from sparse input, which significantly reduces the computational cost, and multiple distillations improve its accuracy. 
The visual comparisons are shown in Fig.~\ref{exp-qualitative}, compared with MHFormer~\cite{li2022mhformer} and PoseFormerV2~\cite{zhao2023poseformerv2}, our SCJD achieves more accurate gesture positions and joint position relationships thanks to the skeleton-prior distillation.  

We also evaluate our method on MPI-INF-3DHP~\cite{mehta2017monocular}. As shown in Table~\ref{table2}, with the same fewer input sequences, our method can best achieve a balance between accuracy and complexity, which proves the effectiveness of our SCJD.

\subsection{Comparisons with Knowledge Distillation Methods}
As shown in Table~\ref{table3}, we re-implemented the general knowledge distillation method and demonstrated that SCJD achieves higher accuracy than other distillation approaches, even with smaller input sequences and lower computational requirements. Unlike traditional knowledge distillation methods, SCJD enables the student network to effectively learn both the spatial dependencies between joints within a pose and the temporal dependencies between frames in a sequence. This is achieved through our carefully designed \textit{Dynamic Joint Spatial Attention Distillation} and \textit{Temporal Consistency Distillation}, resulting in significantly improved accuracy.

\begin{table}
    \setlength{\tabcolsep}{3.5pt}
    \caption{Comparisons with Previous Low-frame-rate Methods on MPI-INF-3DHP. Using ground-truth 2D poses as inputs. }
    \vspace{-8pt}
    \label{table2}
    \centering
    \begin{tabularx}{\columnwidth}{lccrrrr}
        \hline
        Method & Publication & \(f\) & MFLOPs↓ &PCK↑ & AUC↑ & MPJPE↓ \\ 
        \hline
        PoseFormer~\cite{zheng20213d} & ICCV'21 & 9  &\textcolor{blue}{150}& 88.6 & 56.4 & 77.1 \\
        CrossFormer~\cite{hassanin2022crossformer}& arXiv'22&9&163 &89.1&57.5&76.3\\
        PAA~\cite{xue2022boosting} & TIP'22 & 9 &1139.1& 90.3 & 57.8 & 69.4 \\
        MHFormer~\cite{li2022mhformer} & CVPR'22 & 9 &342.9&93.8 & 63.3 & 58.0 \\
        UAU~\cite{einfalt2023uplift} & WACV'23 & 9  &543& 95.4 & 67.6 & 46.9 \\
        DualFormer~\cite{zhou2024tcsvt} &TCSVT’24 &9 &-& 97.8&73.4&40.1\\
        STCFormer~\cite{Tang_2023_CVPR} & CVPR'23 &9&484.1&\textcolor{blue}{98.2}&\textcolor{blue}{81.5}&\textcolor{blue}{28.2}\\
        \rowcolor{mygray}
        \textbf{SCJD (Ours)} && 9  &\textcolor{red}{81.3}& 97.5 & 76.5 & 30.7\\ 
        \rowcolor{mygray}
        \textbf{SCJD (Ours)} && 27 &171.5& \textcolor{red}{98.4} & \textcolor{red}{79.1} & \textcolor{red}{27.0} \\ 
        \hline
    \end{tabularx}
    \vspace{-10pt}
\end{table}

\begin{table}
\setlength{\tabcolsep}{10pt}
    \caption{Comparisons with Traditional KD Methods on Human3.6M. The 2D poses obtained by CPN~\cite{chen2018cascaded} are used as inputs. }
    \vspace{-8pt}
    \label{table3}
    \centering
    \begin{tabularx}{\columnwidth}{lccrr}
        \hline
        Method & Publication & \(f\) &MFLOPs↓ & MPJPE↓ \\
        \hline
        SP~\cite{tung2019similarity} &ICCV'19& 81 & 406.2 & 49.8 \\ 
        DKD~\cite{zhao2022decoupled} &CVPR'22& 81 & 406.2   & 49.3 \\ 
        PKT~\cite{passalis2018learning} &ECCV'18&81 & 406.2  & 48.8 \\ 
        SimKD~\cite{Chen_2022_CVPR} &CVPR'22 & 81 & 406.2 & 47.9 \\
        KDSVD~\cite{lee2018self} &ECCV'18& 81 & 406.2  & 47.7 \\ 
        RKD~\cite{park2019relational} &CVPR'19& 81 & 406.2  & \textcolor{blue}{46.8} \\
        \rowcolor{mygray}
        \textbf{SCJD (Ours)}& & 27  & \textcolor{red}{171.5}& \textcolor{red}{45.2} \\ 
        \hline
    \end{tabularx}
    \vspace{-10pt}
\end{table}

\begin{table}[!t]
\setlength{\tabcolsep}{5pt}
    \caption{Ablation Study on Different Parts on Human3.6M.}
        \vspace{-8pt}
    \label{table4}
    \centering
    \begin{tabularx}{\columnwidth}{ccccrrr}
        \hline
        Sampling & Distillation & \(f\) & RF & Params(M) & MFLOPs↓ & MPJPE↓\\ 
        \hline
        \ding{55} & \ding{55} & 27 & 27 & 2.41 & 135.4 & 51.4 \\ 
        \ding{51} & \ding{55} & 27 & 81 & 2.80 & 171.5 & 48.9 \\ 
        \hline
        \ding{55} & \ding{55} & 81 & 81 & 2.43 & 406.2 & 48.5 \\
        \ding{55} & \ding{51} & 81 & 81 & 2.43 & 406.2 & \textcolor{blue}{46.1} \\
        \hline
        \ding{51} & \ding{51} & 27 & 81 & 2.80 & 171.5 & \textcolor{red}{45.2} \\ 
        \hline
    \end{tabularx}
    \vspace{-10pt}
\end{table}

\begin{table}[!t]
\setlength{\tabcolsep}{7.2pt}
    \caption{Ablation Study on Distillation Components on Human3.6M.}
        \vspace{-8pt}
    \label{table5}
    \centering
    \begin{tabularx}{\columnwidth}{ccccrrr}
        \hline
        Method &\(f\) & \(\mathcal{L}_{reg}\) & \(\mathcal{L}_{emb}\) & \(\mathcal{L}_{am}\) & \(\mathcal{L}_{temp}\) & MPJPE↓ \\ 
        \hline
        w/o $\mathcal{L}_{emd}$  &27 & \ding{51} & \ding{55} & \ding{51} & \ding{51} & \textcolor{blue}{46.5} \\ 
       w/o $\mathcal{L}_{attn}$  &27 & \ding{51} & \ding{51} & \ding{55} & \ding{51} & 48.2 \\ 
        w/o $\mathcal{L}_{temp}$ &27 & \ding{51} & \ding{51} & \ding{51} & \ding{55} & 47.3 \\ 
        full &27 & \ding{51} & \ding{51} & \ding{51} & \ding{51} & \textcolor{red}{45.2} \\ 
        \hline
    \end{tabularx}
    \vspace{-10pt}
\end{table}

\begin{table}[!t]
\setlength{\tabcolsep}{7.5pt}
    \caption{Ablation Study on Parameter Settings on Human3.6M.}
    \vspace{-8pt}
    \label{table6}
    \centering
    \begin{tabularx}{\columnwidth}{ccccrrr}
        \hline
        Model & \(f\) & \(C\) & \(L\)   & Params(M) & MFLOPs↓ & MPJPE↓ \\ 
        \hline
        Teacher & 81 & 32 & 4  & 9.60 & 1358 & 44.3 \\ 
        \hline
        \multirow{6}{*}{Student} & 27 & 16 & 4  & 2.80 & 171.5 & \textcolor{red}{45.2}\\ 
        & 9 & 16 & 4 & 2.79 & 81.3 & \textcolor{blue}{46.3}\\ 
        \cline{2-7}
        & 27 & 8 & 4 & 0.71 & \textcolor{blue}{42.6} & 47.5 \\ 
        & 9 & 8 & 4 & 0.71 & \textcolor{red}{20.1} & 47.8 \\ 
        \cline{2-7}
        & 27 & 16 & 2 & 1.60 & 104.2 & 46.8\\ 
        & 9 & 16 & 2 & 1.60 & 58.8 & 47.2 \\ 
        \hline
    \end{tabularx}
    \vspace{-10pt}
\end{table}

\subsection{Ablation Study}
\textbf{Impact of Different Parts.} In Table~\ref{table4}, we investigate the impact of sequence downsampling and distillation on the student network. RF represents the Receptive Field, which is the length of the sequence before sampling. From Rows 1 and 2, sequence downsampling brings an improvement of 2.5mm MPJPE. We can think that the sparse sequence obtained by sampling is equivalent to expanding the receptive field under the same input length. From Rows 3 and 4,  distillation improves the accuracy by 2.4mm MPJPE, which proves the effectiveness of knowledge transfer. The last row shows that performance is optimal when all modules are included. Compared with the same 27-frame input (Row 1), the performance is significantly improved by 6.2mm MPJPE.

\textbf{Impact of Distillation Components.} In Table~\ref{table5}, we investigate the impact of various distillation components on the student model. Table~\ref{table5} shows that any of the three distillations plays an important role because they all convey corresponding knowledge. Besides, we can find that the impact of \(\mathcal{L}_{am}\) is particularly obvious, which proves that attention distillation combined with skeletal priors better learns joint relationships.

\textbf{The Smaller Model.} In Table~\ref{table6}, we compared the performance of the student and the teacher model. The student model achieves an acceptable loss of accuracy in exchange for significantly fewer Params and FLOPs. To further reduce the student model's size, we respectively reduce the feature dimensions of the joint-based embeddings \(C\) and the number of layers \(L\) in both spatial and temporal transformers. From Rows 4 and 5, we observed that further reducing the embedding dimension reduces the computational cost and maintains respectable accuracy, suggesting that reducing the embedding dimension also preserves feature representation capabilities via distillation. Notably, in Row 5, when \(f\) is 9, 47.8mm MPJPE can be achieved with only 20.1 MFLOPs, which is comparable to the accuracy of MHFormer~\cite{li2022mhformer}, but we are about 17.0 times faster than it. From Rows 6 and 7, reducing depth degrades accuracy, because too few layers will weaken the ability to extract more complex features at different layers.

\section{Conclusion}
This paper proposes SCJD, effectively addresses the limitations of existing 3D HPE methods by combining sparse correlation input sequence downsampling with advanced dynamic joint spatial attention distillation techniques.
By preserving inter-frame correlations, enhancing spatial understanding, and aligning temporal consistency, SCJD significantly reduces computational overhead while maintaining high accuracy. Extensive experiments validate its superior performance, making SCJD a promising solution for efficient and accurate 3D HPE.

\bibliographystyle{IEEEbib}
\bibliography{icme2025references}

    

\clearpage

\appendix

\maketitle

\begin{figure*}[t]
\centering
\centerline{\includegraphics[width=\linewidth]{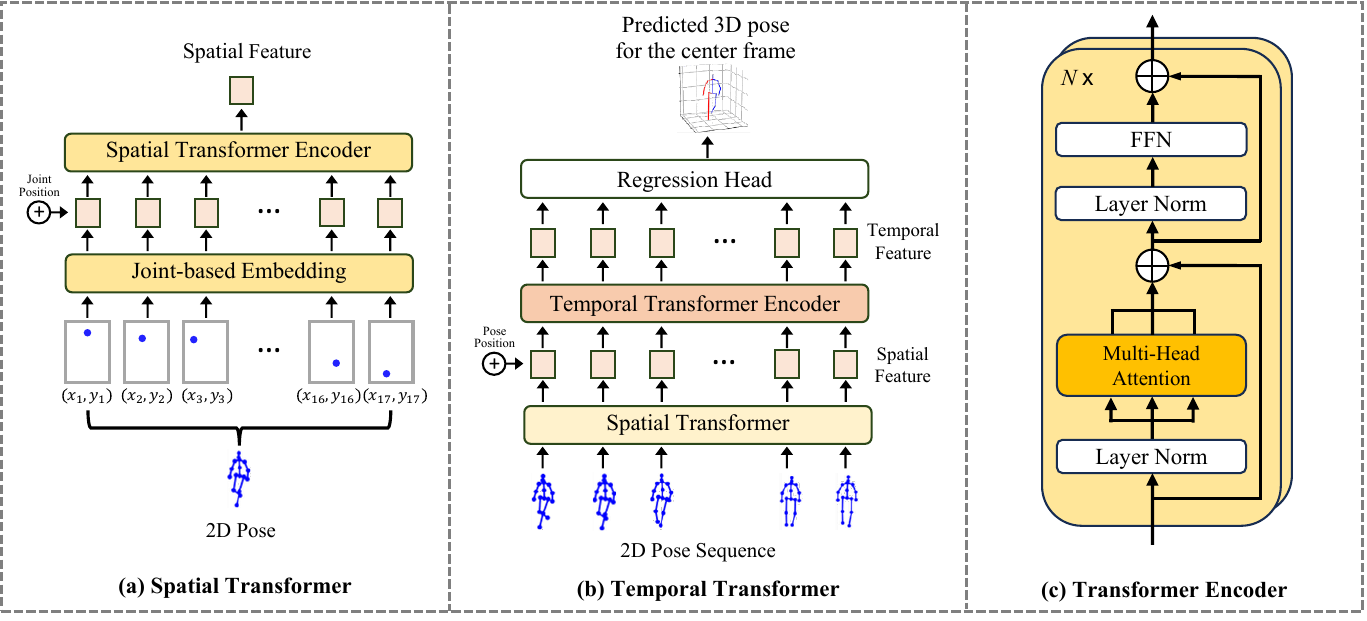}} 
    \caption{Left: details of the spatial transformer. Center: details of the temporal transformer. Right: details of the transformer encoder structure.}
    \label{supp}
\end{figure*}
The supplementary material contains: 1) Additional Details of the Network Structure; 2) More Experimental Details; 3) Qualitative Comparisons on In-the-wild Videos; 4) More Qualitative Comparisons on Human3.6M; 5) Hyperparameter Analysis.

\subsection{Additional Details of the Network Structure}
In our framework, spatial transformers, as shown in Fig.~\ref{supp}(a), are used to model the positional connections among joints in a frame of pose. Building on PoseFormer~\cite{zheng20213d}, for each frame of the input 2D pose sequence, coordinates of all joints of a pose are linearly projected to a higher-dimensional space, and a learnable spatial positional embedding is added to encode the positions of the joints. These features are then passed through a spatial transformer encoder. Then, the output of each frame of the spatial transformer encoder is concatenated as the input of the temporal transformer encoder, as shown in Fig.~\ref{supp}(b). Temporal transformer encoders capture frame-to-frame dependencies across the sequence, and its output is passed through the regression head to obtain the 3D pose for the center frame.

The spatial transformer encoder and temporal transformer encoder in our framework both adopt the standard transformer encoder architecture~\cite{vaswani2017attention}, as illustrated in Fig.~\ref{supp}(c). This architecture consists of \textit{N} identical layers, each comprising two main components: a Multi-Head Self-Attention (MHSA) mechanism and a Feed Forward Network (FFN). Layer Normalization (Layer Norm) is applied before the MHSA and FFN to stabilize training, while residual connections are employed around these sublayers to improve gradient flow and facilitate efficient learning. The MHSA module enables the encoder to capture complex dependencies within the input sequences, while the FFN refines the intermediate representations. 

\begin{figure*}[t]
    \centering
\centerline{\includegraphics[width=\linewidth]{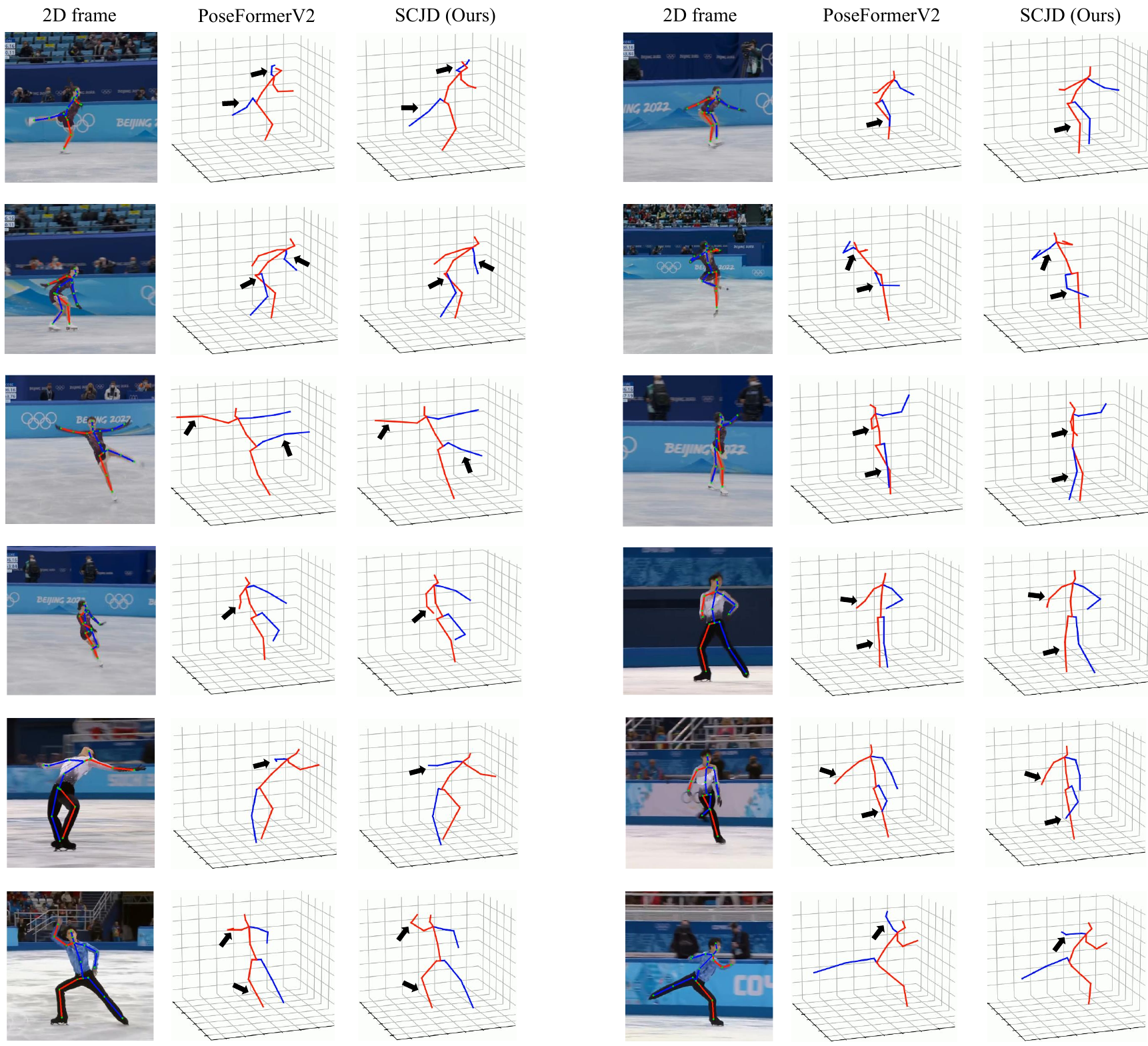}}
    \caption{Qualitative comparisons between ours and PoseFormerV2~\cite{zhao2023poseformerv2} on in-the-wild videos. The same visualization setting as PoseFormerV2~\cite{zhao2023poseformerv2} is adopted, the 2D pose is obtained by the detector HRNet~\cite{HRnet}, and the number of input frames of the model is \textit{f}=27. The black arrows highlight locations where our method clearly has better results.}
    \label{wild}
\end{figure*}
\section{More Experimental Details}
The datasets used in our experiment are Human3.6M~\cite{ionescu2013human3} and MPI-INF-3DHP~\cite{mehta2017monocular}, which are the most widely used in 3D human pose estimation. 

\textbf{Human3.6M} is the most widely used public indoor dataset for 3D human pose estimation. Eleven professional actors performed 15 actions including sitting, walking, discussing, photoing, sitting down, etc. Human3.6M consists of 3.6 million video frames with 3D ground truth annotations captured by a precise marker-based motion capture system. We adopt the same experimental setup as the previous work~\cite{eccv_repose, Peng_2024_CVPR, Tang_2023_CVPR, zhang2022mixste, li2022mhformer}: the model is trained on five sections (Subjects S1, S5, S6, S7, S8), and tested on two sections (Subjects S9 and S11). In common with existing 2D to 3D lifting work, we use 2D poses from CPN~\cite{chen2018cascaded} during student network training and evaluation.

\textbf{MPI-INF-3DHP} is smaller but more challenging than Human3.6M. In addition to indoor collection, it also collects data outdoors, containing different themes and actions. We use ground truth 2D poses in our experiments on MPI-INF-3DHP for comparison with existing work~\cite{Tang_2023_CVPR, zhou2024tcsvt, li2022mhformer, xue2022boosting, hassanin2022crossformer}.

\textbf{Evaluation Metrics.} For Human3.6M, we use the most common evaluation metric Mean Per Joint Position Error (MPJPE), the average euclidean distance in millimeters between the predicted and the ground-truth 3D joint coordinates, to evaluate our estimation performance. For MPI-INF-3DHP, we use MPJPE, Percent Correct Keypoints (PCK) within a range of 150 mm, and Area Under the Curve (AUC) with a threshold range of 5-150 mm. To evaluate the efficiency of the model, we use FLOating-Point operations (FLOPs) to measure the computational cost.

\textbf{Implementation Details.} The input sequence length of the teacher model is \(f_T = 81\). For the student model, we chose three different frame sequence lengths, i.e., \(f_S= 3\), \(f_S= 9\), and \(f_S= 27\). Pose level flipping was applied as a data augmentation in training and testing~\cite{pavllo20193D}. We trained our model using the Adam optimizer for 50 epochs with a weight decay of 0.1. We adopt an exponential learning rate decay schedule with an initial learning rate of 1e-3 and a decay factor of 0.98 for each epoch. An NVIDIA GeForce RTX 4090 GPU is used for training with the batch size set to 1024. The details about model parameters of teacher network and student network with results are discussed in the ablation studies (Sec. IV-C of the article: The Smaller Model).

\begin{figure*}[t]
    \centering
    \centerline{\includegraphics[width=\linewidth]{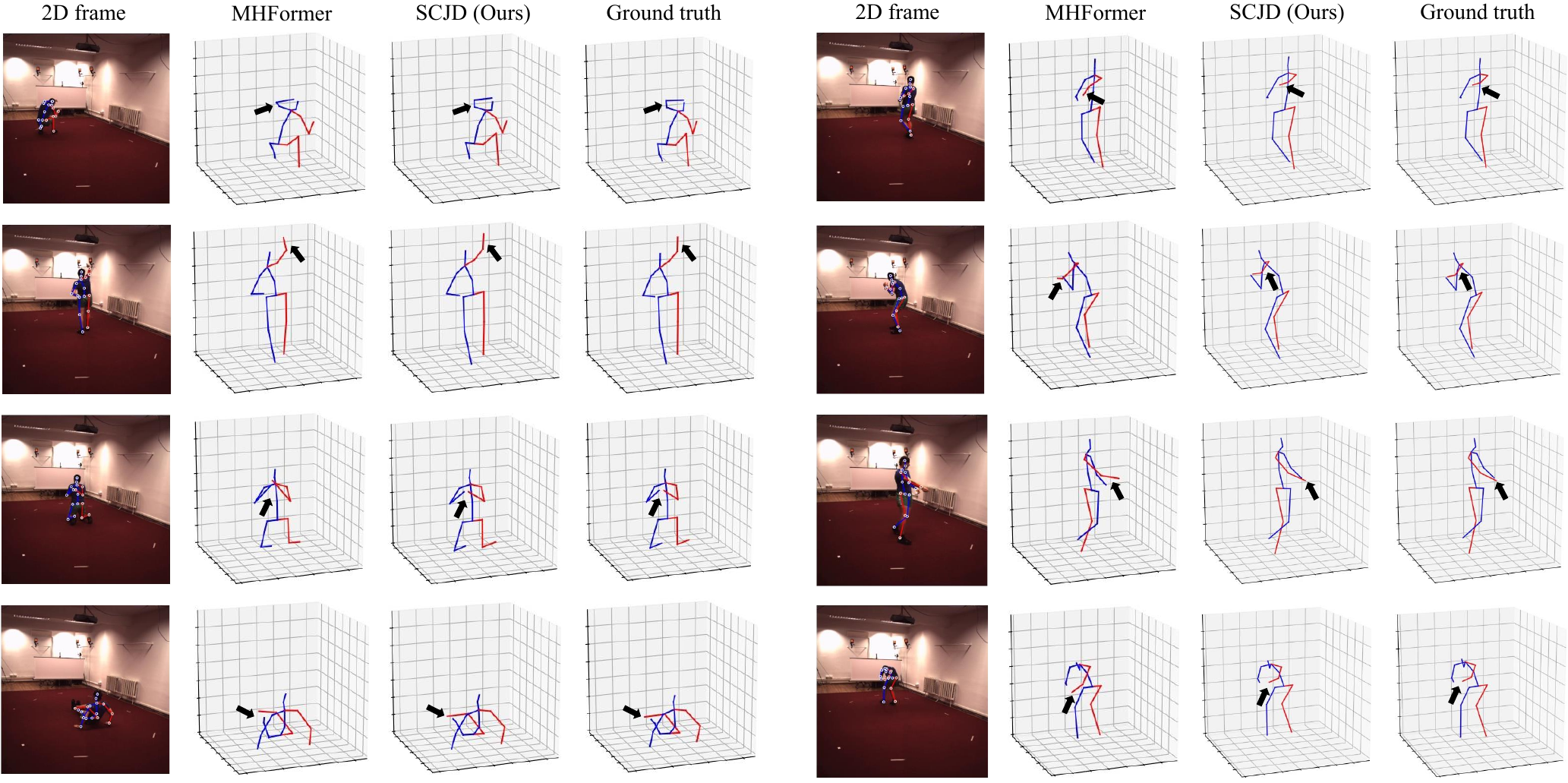}}
    \caption{Comparison of visualization results between ours and MHFormer~\cite{li2022mhformer} on Human3.6M test set S11 with the \textit{Photo}, \textit{Sitting Down}, and \textit{Walkdog} actions. The 2D pose is obtained by the detector CPN~\cite{chen2018cascaded}, and the number of input frames of the model is \textit{f}=27. The black arrows highlight locations where our method clearly has better results. }
    \label{hm36}
\end{figure*}

\subsection{Qualitative Comparisons on In-the-wild Videos}
To further validate the generalization ability of our model, we collected some wild videos as additional real-world tests. In our videos, figure skating is a sport with large movements, complex postures, and fast speeds, especially with frequent occlusions between body joints. As shown in Fig.\ref{wild}, our method shows excellent accuracy in most frames of the wild videos. The black arrows point out the locations where our method is significantly more accurate than the state-of-the-art efficient pose estimation method PoseFormerV2~\cite{zhao2023poseformerv2}. Our results are especially better in fast-moving poses, which proves that the student network can learn richer contextual information from the teacher network. In addition, our joint position relationship is more reasonable, thanks to the adjacent joint attention distillation. 

\subsection{More Qualitative Comparisons on Human3.6M}
In this section, we supplement and present more qualitative results of our SCJD. Photo, Sitting Down, and Walkdog are more challenging actions in the Human3.6M test set. Fig.~\ref{hm36} shows a visual comparison of 3D pose estimation results between our SCJD and the representative method MHFormer~\cite{li2022mhformer}. The black arrow highlights locations where we clearly has better results compared to MHFormer~\cite{li2022mhformer}. It can be intuitively seen that our SCJD achieves more accurate hand and leg posture positions and joint position relationships. 

\subsection{Hyperparameter Analysis} 
We have further analyzed the hyperparameters of the loss function (Eq. 8), with the corresponding results summarized in Table~\ref{re-tab-1}. Specifically, we examine the weights \(\alpha\), \(\beta\), and \(\gamma\), which correspond to the individual loss components.
The results indicate that our method achieves the best performance when \(\alpha = \beta = 1\) and \(\gamma = 0.01\). We attribute this to the fact that these settings effectively scale the losses to the same order of magnitude, thereby facilitating more stable and balanced optimization. 

\begin{table}[h]
\centering
\setlength{\tabcolsep}{18pt}
\caption{Ablation Study on Parameter Setting of Loss Function.}
\label{re-tab-1}
    \resizebox{0.35\textwidth}{!}{
    \large
    \begin{tabular}{cccc}
        \hline
       $\alpha$ & $\beta$ & $\gamma$ &  MPJPE↓ \\
        \hline
        1 & 1 & 0.01 & \textbf{45.2} \\
        1 & 1 & 0.1 & 45.3 \\
        1 & 1 & 1 & 45.3 \\
        0.01 & 0.01 & 1 & 45.7 \\
        0.1 & 0.1 & 1 & 45.5 \\
        \hline
    \end{tabular}
    }
\end{table}


\end{document}